\documentclass{isrpaper}

\usepackage{tabularx}
\newcolumntype{C}{>{\centering\arraybackslash}X}
\usepackage{makecell}
\usepackage{threeparttable}
\usepackage{graphicx}
\usepackage{float}
\usepackage{amsmath,amssymb,amsfonts}
\usepackage{textcomp}
\usepackage{xcolor}
\usepackage{booktabs}
\usepackage{ragged2e}  % allows \justifying

\usepackage{orcidlink}

\usepackage[
    backend=biber,
    style=authoryear,
    natbib=true,
    doi=false,
    url=false,
    isbn=false,
    maxbibnames=99,
    maxcitenames=2,
    uniquelist=false,
    giveninits=true,    % "Cotton, R. J." instead of "Cotton, R. James"
    terseinits=true,    % "R.J." instead of "R. J."
]{biblatex}

\AtEveryBibitem{%
   \clearfield{Title}%
   \clearfield{eprint}%
   \clearfield{note}%
}

\renewcommand{\cite}[1]{\unskip~\parencite{#1}}

\usepackage{etoolbox}

\AtBeginEnvironment{tabular}{\footnotesize}
\AtBeginEnvironment{tabularx}{\footnotesize}
\usepackage[font=small,labelfont=bf,labelsep=period,
            justification=justified,format=plain,singlelinecheck=false]{caption}
\title{Markerless Motion Capture for Biomechanical Whole-Body Kinematic Estimation in Infants}

\author{
Divya Joshi\,\orcidlink{0000-0001-6397-0037}\,$^{1,2}$,
J.D. Peiffer\,\orcidlink{0000-0003-2382-8065}\,$^{1,3}$,
Colleen Peyton\,\orcidlink{0000-0003-4753-7246}\,$^{2,4}$,
R. James Cotton\,\orcidlink{0000-0001-5714-1400}\,$^{1,5}$}

\affiliations{
$^{1}$Center for Bionic Medicine, Shirley Ryan AbilityLab, Chicago, IL \\
$^{2}$Dept. of Physical Therapy and Human Movement Science,\\Northwestern University, Chicago, IL \\
$^{3}$Dept. of Biomedical Engineering,\\Northwestern University, Evanston, IL \\
$^{4}$Dept. of Pediatrics,\\Northwestern University, Chicago, IL \\
$^{5}$Dept. of Physical Medicine and Rehabilitation,\\Northwestern University, Chicago, IL}

\paperdate{May 2026}

\correspondence{rcotton@sralab.org}

\paperabstract{Early identification of motor impairment in infancy relies on expert visual assessment of spontaneous movement, motivating the development of automated, objective alternatives. One promising approach is using computer vision, which benefits from high quality pose estimation from video. In this study, we systematically evaluated three state-of-the-art pose estimation frameworks (MeTRAbs-ACAE, SAM 3D Body, and Sapiens) on 100 videos over 13 sessions of 8 infants recorded with a multi-view markerless motion capture system. We quantified keypoint detection accuracy using reprojection error, geometric consistency, and Procrustes-aligned 3D position error, and demonstrated proof-of-concept for fitting an inverse kinematic framework to infant data. While Sapiens achieved the lowest reprojection error and highest geometric consistency of the methods evaluated (22.8 pixels and 0.82, respectively), SAM 3D Body provided the most comprehensive 3D information for kinematic reconstruction with Procrustes-aligned position errors of 19 to 28 mm. We demonstrate in a case comparison example that biomechanical models fit to SAM 3D estimates distinguish representative movement patterns in infants related to motor development, as identified by a clinical expert. Together, these findings highlight both the promise and current limitations of 3D pose estimation for infant biomechanics and establish preliminary groundwork for scalable, video-based assessment of early motor development.}

\begin{document}
\maketitle

\section{Introduction}

Observation of spontaneous infant movements is an effective method for the early identification of infants at risk for cerebral palsy and other movement disorders\cite{Novak2017,Herskind2015EarlyPalsy}. Clinical assessments such as the Prechtl General Movement Assessment (GMA)\cite{Bosanquet2013AChildren,Einspeiler2005}, the Test of Infant Motor Performance (TIMP)\cite{Spittle2008ALife}, and the Baby Observational Selective Control AppRaisal (BabyOSCAR)\cite{Peyton2024BabyYears,SukalMoulton2024BabyPalsy,Barbosa2024BabyPerformance} evaluate characteristic movement patterns in three month old infants that reflect neuro-motor integrity and help predict developmental trajectories. However, these assessments require extensive rater training, are time-consuming to score, and are susceptible to inter-rater variability\cite{Noble2012NeonatalReview,Bosanquet2013AChildren}, which prevents many infants from receiving these assessments even when indicated and drives a need for automated methods to quantify infant movement.

Markerless motion capture offers a non-invasive and scalable alternative for estimating infant movement from video recordings\cite{Hesse2020LearningSequences,Chambers2020ComputerRisk,Segado2025Data-DrivenFeatures}. This information can be used to derive kinematic measures and canonical movement patterns\cite{Peiffer2024FusingKinematics,Peiffer2025PortableSmartphone,Cotton2024DifferentiableCapture}, analogous to those evaluated in manual clinical assessments, providing insight into early motor development.

A limited number of studies have evaluated the performance of various pose estimation models in infant datasets, including OpenPose, MediaPipe, AlphaPose, HRNet and ViTPose\cite{GamaAutomaticMethods,Ali2024APalsy,Jahn2025ComparisonEstimation,Groos2022DevelopmentRisk}. While several of these methods — such as ViTPose and HRNet — demonstrate strong detection accuracy\cite{GamaAutomaticMethods,Jahn2025ComparisonEstimation}, they present significant limitations for downstream kinematic analysis. In particular, these algorithms output a relatively sparse set of anatomical keypoints (typically ranging from 17 to 33), which is insufficient for calculating whole-body kinematics. Furthermore, all of these approaches provide only 2D pose estimates, preventing comprehensive quantification of 3D movement in space. Finally, none of these approaches estimate finger or toe keypoints, despite evidence that individuated distal joint movements are critical indicators of healthy neuro-motor development\cite{Cahill-Rowley2014EtiologyPalsy}.

More recent pose estimation frameworks have begun to address these limitations. MeTRAbs estimates 3D poses using volumetric heatmaps \cite{Sarandi2020MeTRAbs:Estimation,Mohammed2023StrengthsTherapy} and produces a dense set of 580 keypoints, enabling more detailed kinematic analysis\cite{Peiffer2025PortableSmartphone, JamesCotton2023OptimizingData, Cotton2024DifferentiableCapture}. Another approach, SAM 3D Body, generates a full parametric mesh of the body, including the hands and feet\cite{YangSAMRecovery}. Finally, Sapiens, which was specifically designed to capture high-resolution features\cite{Khirodkar2024Sapiens:Humans.}, can estimate 308 keypoints, including detailed finger joints and facial points. However, it lacks dense keypoint coverage over the torso, which is essential for biomechanical analysis, and does not directly produce 3D keypoints, though it does offer depth and body segmentation outputs. A key challenge in computer vision is the lack of a standardized keypoint definition, resulting in substantial variation in keypoint locations and densities across methods\cite{Sarandi2020MeTRAbs:Estimation}. Accordingly, this work evaluates the 3D accuracy and consistency of different approaches rather than the sufficiency of any particular keypoint set\cite{JamesCotton2023OptimizingData}.

Although all three methods show promising results in benchmark adult datasets\cite{Sarandi2020MeTRAbs:Estimation,YangSAMRecovery,Khirodkar2024Sapiens:Humans.}, their applicability to infant datasets is still unknown. This study aims to systematically evaluate these three algorithms using videos of infant spontaneous movements recorded with a Multi-view Markerless Motion Capture (MMMC) system\cite{Cotton2023MarkerlessPipeline}. Our primary goal was to compare the 3D accuracy of each method in detecting anatomical keypoints in one- to four-month-old infants. The use of an MMMC system allows for reconstructions from multiple synchronized camera views, providing a high quality reference for comparison to single-camera based estimations without requiring manual keypoint annotation or marker-based validation, which is impractical in infant populations. Additionally, we explored whether our previously developed MuJoCo-based inverse kinematic framework \cite{Cotton2026MonocularModels} could be fit to 3D keypoints detected from a monocular, clinical video recording and reproduce joint angles during representative infant movements as identified by an expert clinician.

\section{Methods}

\subsection{Participants}

The study cohort included seven infants aged 8 to 16 weeks. All infants were born at term age and had no health complications. The parents and/or guardians of all participants provided informed, written consent to participate in this study, which was approved by the institutional review board of Northwestern University.

\subsection{Video Acquisition and Preprocessing}

\begin{figure}[!htbp]
    \centering
    \includegraphics[width=\columnwidth]{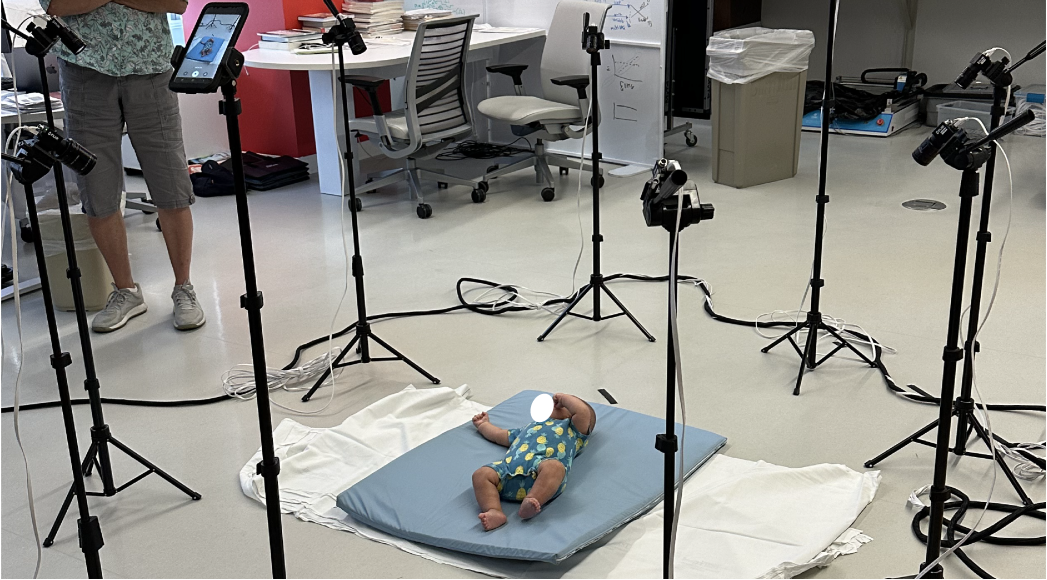}
    \caption{\textbf{The MMMC experimental setup.} The infant is placed supine on a mat in the center of 8 tripods with cameras pointing at the mat.}
    \label{fig1}
\end{figure}

Each infant was dressed in a solid-colored onesie and placed in the center of a mat. Video recordings of the infant were acquired using a MMMC system consisting of 8 FLIR BlackFly S GigE cameras equipped with f/1.4 lenses (nominal 6 mm focal length, with focus adjusted as needed) that captured synchronized RGB video at 29 frames per second. As shown in Fig. \ref{fig1}, the cameras were arranged on tripods surrounding the infant, such that the infant was in the field of view for each camera \cite{Cotton2023MarkerlessPipeline, Cotton2023ImprovedEstimation}. For each session, 8 to 10 trials of 50 to 60 seconds each were recorded while the infant was spontaneously moving.

All downstream processing and analysis was done in PosePipe\cite{Cotton2022PosePipe:Research}. For each session, camera calibration was performed using videos of a moving checkerboard or ChAruco board (7×5 grid), with square sizes of 38 mm. Intrinsic and extrinsic camera parameters were estimated using the Anipose library\cite{Karashchuk2021Anipose:Estimation}. The resulting calibrated parameters define a projection function, denoted as $\Pi_i$ for each camera \textit{i}, which projects points from 3D space to the 2D image plane of the corresponding camera. EasyMocap\cite{EasyMoCap} was used to identify the infant in each recording session. Following EasyMocap processing and manual annotations, bounding boxes enclosing the infant were obtained for each camera view in every video.

\subsection{Evaluation Dataset}

The dataset evaluated in this study comprised the following:

\begin{itemize}
    \item \textbf{Participants:} 8 infants, 5 to 16 weeks post-menstrual age
    \item \textbf{Sessions:} 13 unique sessions across all participants
    \item \textbf{Cameras:} 8 synchronized, calibrated views per session
    \item \textbf{Recordings:} 124 multi-view recordings (992 total videos)
    \item \textbf{Total frames:} 216,011 synchronized multi-view frames
\end{itemize}

Each session provided multi-view coverage from 8 cameras, enabling robust 3D reconstruction via triangulation and reprojection optimization.

\subsection{Keypoint Detection}

For each video across all trials and camera views, three keypoint detection algorithms were applied in a top-down manner using the previously identified bounding boxes.

\subsubsection{MeTRAbs} We employed MeTRAbs-ACAE\cite{Sarandi2020MeTRAbs:Estimation} to estimate 2D and 3D virtual marker locations from RGB video. Although MeTRAbs produces a dense set of 580 keypoints, we restricted our analysis to the 87 keypoints defined by the MoVi dataset\cite{Ghorbani2021MoVi:Dataset}, a subset which we previously demonstrated is sufficient for accurate whole-body kinematic analysis, excluding the fingers\cite{Cotton2024DifferentiableCapture}. Notably, the MoVi keypoint set includes a rich distribution of landmarks across the pelvis and torso that are not available in more commonly used pose estimation datasets, enabling more detailed characterization of core body motion. Here, we refer to this keypoint set as \textit{MeTRAbs (MoVi 87)}.

\subsubsection{SAM 3D Body} We used SAM 3D Body for keypoint detection\cite{YangSAMRecovery}, which we ported to an Equinox/JAX-based implementation\cite{Cotton2026MonocularModels}. This model outputs a full 3D parametric mesh of the body by predicting the parameters of the Momentum Human Rig (MHR) from single images. MHR provides 127 joints along a kinematic tree that decouples the internal skeleton and external pose, mesh vertices at multiple resolutions (up to 18,439 for the standard level of detail, used here), and learned regressors that predict 308 keypoint locations corresponding to the Goliath keypoint set \cite{Ferguson2025MHR:Rig, Khirodkar2024Sapiens:Humans., Cotton2026MonocularModels}. The standard inference pipeline reports a 70-keypoint subset consisting of core anatomical landmarks most closely associated with the kinematic tree, which we use downstream in this work and refer to as \textit{SAM 3D Body (MHR 70)}. Additionally, we evaluate the 127 joints along the kinematic tree, which we refer to here as \textit{SAM 3D Body (MHR 127)}. Finally, we derived a mapping to find a combined set of SAM 3D Body joint, mesh vertex, and regressed keypoint outputs that best corresponded to the 87 MoVi landmark sites defined in our previous biomechanical model \cite{Cotton2024DifferentiableCapture,Cotton2026MonocularModels}. In this work, we refer to this keypoint set as \textit{SAM 3D Body (MoVi 87)}. For each of the three SAM 3D Body keypoint sets, we projected the predicted 3D keypoints onto the corresponding camera’s 2D image plane using the focal length predicted by SAM 3D Body.

\subsubsection{Sapiens} Finally, we used the Sapiens Pose 1b model\cite{Khirodkar2024Sapiens:Humans.}, also ported to an Equinox/JAX-based implementation. Sapiens outputs the Goliath keypoint set, consisting of 308 2D landmarks distributed across the body, hands, and face \cite{Khirodkar2024Sapiens:Humans.}. In this work, we refer to this keypoint set as \textit{Sapiens (Goliath 308)}.

We ported SAM 3D Body and Sapiens in Equinox/JAX to integrate with our existing biomechanical fitting pipeline, which uses the JAX-compatible MuJoCo-MJX high-performance physics engine\cite{Todorov2012MuJoCo:Control,Cotton2024DifferentiableCapture}.

\begin{figure*}[!tbp]
    \centering
    \includegraphics[width=\textwidth]{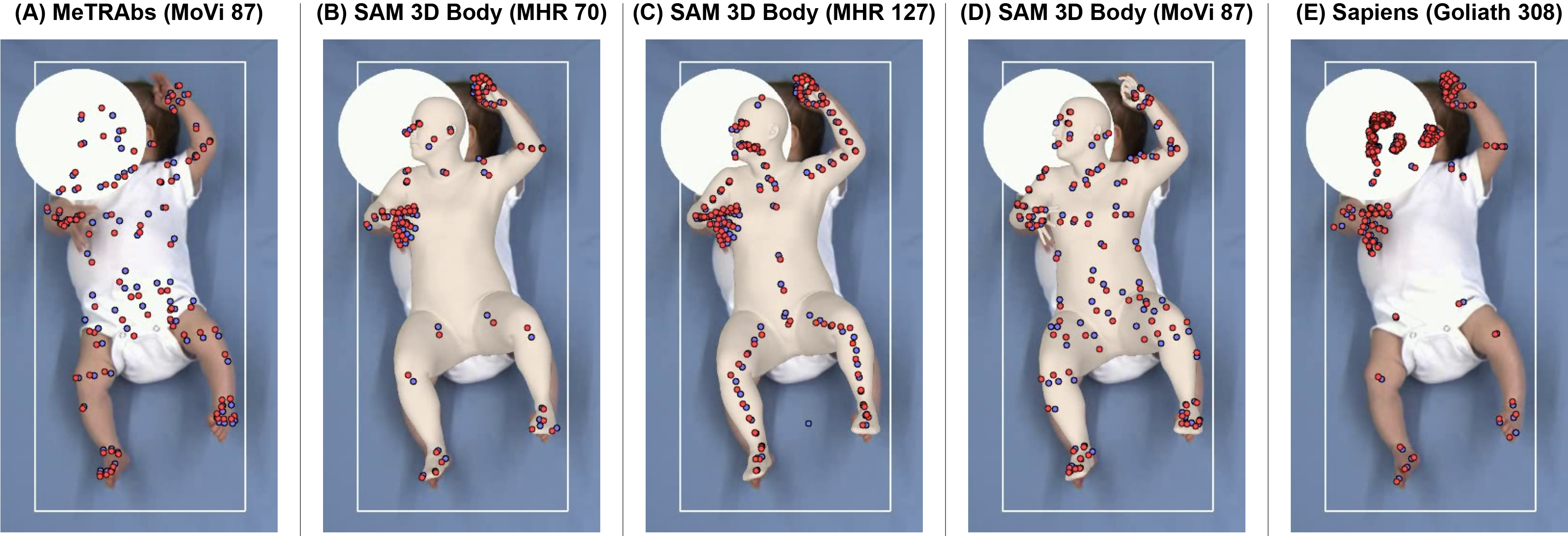}
    \caption{\textbf{Example of detected and reprojected 2D keypoints across methods.} Detected keypoints are shown in red and reprojected keypoints are shown in blue. For (B), (C) and (D), fitted parametric mesh is shown in white. (A) MeTRAbs with the MoVi 87 keypoint set, (B) SAM 3D Body with the MHR 70 keypoint set, (C) SAM 3D Body with the MHR 127 keypoint set, (D) SAM 3D Body with the MoVi 87 keypoint set, and (E) Sapiens with the Goliath 308 keypoint set.}
    \label{fig2}
\end{figure*}

\subsection{Triangulation}
After keypoint detection, we applied a robust triangulation method\cite{Roy2022OnEstimation, Cotton2023ImprovedEstimation} to reconstruct 3D keypoints from the 2D detections across all camera views. Reconstructed 3D keypoints from multi-view triangulation served as a reference for validation, as dense physical marker placement is not feasible in infants. The robust triangulation approach downweights keypoints with low confidence or inconsistent measurements, improving the accuracy of the 3D reconstruction. The resulting 3D points were then reprojected onto each camera’s 2D image plane using the previously defined function $\Pi_i$.

\subsection{Performance Metrics}

\subsubsection{Reprojection Error} For each keypoint and camera, the reprojection error was calculated as the pixel-wise distance between the detected 2D keypoint and its reprojected counterpart, and these errors were averaged across all frames and cameras in the trial. Here, we report reprojection error averaged across keypoints as well as per keypoint reprojection error.

\subsubsection{Geometric Consistency} Geometric consistency (GC$_{d, \lambda}$) is defined as the fraction of keypoints whose reprojection error falls below a threshold \textit{d} pixels, conditioned on the keypoint confidence interval exceeding \textit{$\lambda$}. Here, we report GC$_{5, 0.5}$ and GC$_{10, 0.5}$ (referred to as GC$_{5}$ and GC$_{10}$, respectively, from this point onward) to quantify geometric consistency across keypoints \cite{Cotton2023ImprovedEstimation}.

\subsubsection{Position Error} We compared the monocular 3D estimates against the 3D multi-view triangulated 3D estimates. First, we performed Procrustes alignment on the keypoints estimated from the monocular recording taken by the front-facing camera, which is the camera position that mirrors standard clinical practice for movement assessments, with the multi-view triangulated keypoints as the reference. Then, position error was calculated as the Euclidean distance between the Procrustes-aligned monocular 3D keypoints and the 3D multi-view triangulated keypoints.

For average performance metrics, group differences across methods were first assessed using a Kruskal–Wallis test. When the omnibus test was significant, post hoc pairwise comparisons were performed using Mann–Whitney U tests with Holm correction for multiple comparisons.

\section{Results}

\subsection{Keypoint Detection — Qualitative}

We first visually examined the detected and reprojected 2D keypoints for all three algorithms. An example is shown in Fig. \ref{fig2}. MeTRAbs and SAM 3D Body (MoVi 87) distributes keypoints broadly across the body, with particularly good coverage across the torso. In contrast, SAM 3D Body (MHR 70) and Sapiens concentrate more keypoints at the hands, but with sparser coverage on the limbs and torso.

We observe that the overlap between reprojected and detected keypoints is comparable across methods. For limb keypoints, particularly at hinge joints such as the elbow and knee, SAM 3D Body (MoVi 87) shows better alignment with the underlying anatomical landmarks. In contrast, MeTRAbs exhibits greater discrepancies between reprojected and detected keypoints at these hinge joints. The methods that include detailed hand and face landmarks (SAM 3D Body (MHR 70), SAM 3D Body (MHR 127) and Sapiens) accurately identify detailed hand landmarks. Sapiens further distinguishes itself through its fine-grained facial representation, capturing 274 facial keypoints, although facial landmarks are not the focus of our biomechanical analysis.

\subsection{Keypoint Detection — Performance Metrics}

Table \ref{tab1} summarizes reprojection error, GC$_{5}$, GC$_{10}$, and position error averaged across all keypoints, excluding facial landmarks, for the five evaluated keypoint detection methods. Overall differences across methods were observed for reprojection error (\textit{p}=6.13e-5), GC$_{5}$ (\textit{p}=3.20e-4), GC$_{10}$ (\textit{p}=7.80e-4), and position error (\textit{p}=3.52e-5). Sapiens had roughly two-thirds the reprojection error of all other methods (all adjusted \textit{p}$<$0.01), and approximately twice the GC$_{5}$ of MeTRAbs and all SAM 3D Body methods (all adjusted \textit{p}$<$0.05), as well as a higher GC$_{10}$ than all SAM 3D Body methods (all adjusted \textit{p}$<$0.05). For position error, SAM 3D Body (MHR 70) and SAM 3D Body (MoVi 87) were both significantly lower than MeTRAbs (both adjusted \textit{p}$<$0.01) and SAM 3D Body (MHR 127) (both adjusted \textit{p}$<$0.05). No other pairwise differences were significant across any metric.

\begin{table*}[!tbp]
\caption{Performance metrics across keypoint detection methods.}
\begin{center}
\begin{threeparttable}
\begin{tabularx}{\textwidth}{@{}lCCCCC@{}}
\toprule
& \multicolumn{5}{c}{\textbf{Keypoint Detection Method}} \\
\cmidrule(l){2-6}
& \textbf{MeTRAbs}
& \multicolumn{3}{c}{\textbf{SAM 3D Body}}
& \textbf{Sapiens} \\
& (MoVi 87)
& (MHR 70) & (MHR 127) & (MoVi 87)
& (Goliath 308) \\
\midrule
Mean reprojection error\tnote{*} (pixels) & 33.8 ± 10.8 & 33.8 ± 13.1 & 33.2 ± 12.8 & 32.7 ± 11.2 & 22.8 ± 4.2 \\
GC$_{5}$\tnote{*}                         & 0.25 ± 0.1  & 0.29 ± 0.1  & 0.32 ± 0.1  & 0.29 ± 0.1  & 0.50 ± 0.2 \\
GC$_{10}$\tnote{*}                        & 0.60 ± 0.2  & 0.65 ± 0.2  & 0.68 ± 0.1  & 0.65 ± 0.1  & 0.82 ± 0.1 \\
Mean position error (mm)\tnote{*}         & 29.3 ± 4.9  & 19.5 ± 5.4  & 22.2 ± 4.7  & 27.6 ± 3.9  & N/A        \\
\bottomrule
\end{tabularx}

\begin{tablenotes}[flushleft]
\footnotesize
\item[*] Overall difference between methods was significant (\textit{p}$<$0.05), as detected by a Kruskal-Wallis test. Mean reprojection error, GC$_{5}$, and GC$_{10}$ were computed over non-facial keypoints only. Mean position error is reported only for methods that include monocular 3D estimates.
\end{tablenotes}
\end{threeparttable}
\end{center}

\label{tab1}
\end{table*}

\begin{figure*}[!tbp]
    \centering
    \includegraphics[width=\textwidth]{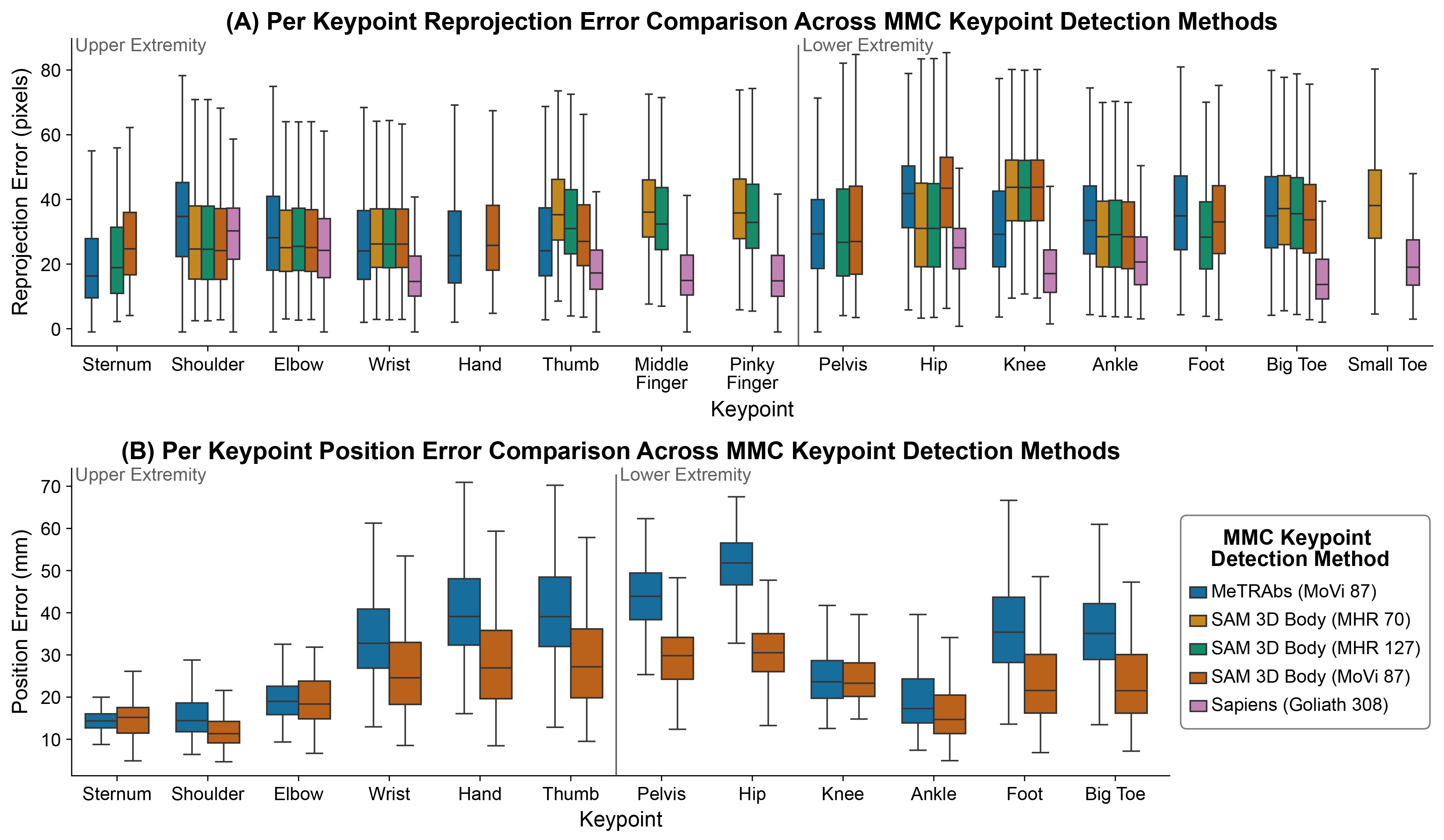}
    \caption{\textbf{Per keypoint reprojection error and position error across keypoint detection methods}. MeTRAbs is shown in blue, SAM 3D Body (MHR 70) is shown in orange, SAM 3D Body (MHR 127) is shown in green, SAM 3D Body (MoVi 87) is shown in red, and Sapiens is shown in pink. Boxplots represent the median, interquartile ranges, and minimums/maximums across participants and trials.}
    \label{fig3}
\end{figure*}

Next, we examined the reprojection error for a selected set of  keypoints (Fig. \ref{fig3}A). At the sternum, MeTRAbs exhibited a lower reprojection error (about 15 pixels) than the SAM 3D Body (MoVi 87) and SAM 3D Body (about 20 pixels), but all three were similar at the pelvis (about 30 pixels). In the upper extremity, all five approaches showed comparable reprojection error at the shoulder (approximately 20 to 30 pixels), but performance diverged at more distal joints. At the wrist and the thumb, Sapiens achieved a reprojection error of 15 to 20 pixels, about half that of MeTRAbs and all SAM 3D Body methods, which ranged from 30 to 40 pixels. Although the MoVi dataset does not include keypoints for the remaining fingers, the same trend of lower reprojection error for Sapiens relative to SAM 3D Body (MHR 70) and SAM 3D Body (MHR 127) was observed across those landmarks. In the lower extremity, Sapiens showed reprojection errors of about 15 to 20 pixels, about half those of the other methods (30 to 40 pixels). The other methods performed similarly across the lower extremity, except at the knee where MeTRAbs achieved a lower reprojection error at about 30 pixels, compared to the three SAM 3D Body methods with reprojection errors at the knee of about 45 pixels.

Fig. \ref{fig3}B shows the position error for a selected set of keypoints for the MeTRAbs and SAM 3D Body (MoVi 87) methods, as the MoVi 87 keypoint set is of interest for downstream biomechanical analysis. In the proximal upper extremity (sternum to elbow), the two methods exhibited similar position errors (15 to 20 mm); however, their performance diverged at distal upper-extremity joints. At the wrist, hand, and thumb, SAM 3D Body (MoVi 87) achieved position errors of approximately 25 mm, compared with 30 to 40 mm for MeTRAbs. In the lower extremity, SAM 3D Body showed position errors that were comparable to MeTRAbs at the knee and ankle (15 to 25 mm), but substantially lower at the pelvis, hip, foot, and big toe (20 to 30 mm vs. 30 to 50 mm).

Finally, because Sapiens achieved the best performance in terms of reprojection error and geometric consistency, but does not provide monocular 3D estimates, we used multi-view triangulated Sapiens 3D keypoints as an additional high-quality reference for position error computation. These Sapiens 3D estimates were compared against the monocular 3D estimates produced by SAM 3D Body (MHR 70), both across the full body and for hand keypoints only. The mean position error across all keypoints was 23.8 $\pm$ 7.6 mm, comparable to the position errors reported in Table \ref{tab1}, whereas the mean position error restricted to hand keypoints was substantially lower at 6.80 $\pm$ 1.3 mm. This comparison could not be performed for MeTRAbs or SAM 3D Body (MoVi 87), as these keypoints are different than the Goliath keypoint definitions.

\section{Kinematic Reconstruction in a Clinical Case Example}

\subsection{Methods}

A limitation of mesh representations such as MHR are they do not parameterize motion according to clinical and biomechanical conventions. Our previous work has established an inverse kinematics (IK) framework for fitting a biomechanical model to video-based pose estimates \cite{Cotton2026MonocularModels}, built on the full body MuJoCo model, a GPU-accelerated physics simulation commonly used for musculoskeletal modeling \cite{Todorov2012MuJoCo:Control}. We refer to our previous work for further details on the IK approach  \cite{Cotton2026MonocularModels}.

In this work, we applied the IK framework to keypoints obtained using the SAM 3D Body (MoVi 87) approach, in a monocular video from our clinical dataset of infant recordings. To better represent infant anatomical proportions, we introduced separate scale parameters for the head, torso, upper leg, lower leg, foot, upper arm, forearm, and hand. We also include an overall scaling parameter, that can capture anatomical landmarks that are not captured in the previously mentioned set. An expert clinician then identified instances of representative knee movements corresponding to those scored in standard clinical assessments \cite{Peyton2024BabyYears,SukalMoulton2024BabyPalsy,Barbosa2024BabyPerformance}, and we compared the joint angles estimated by the IK reconstruction to the patterns identified by the clinical rating.

\subsection{Results}

\begin{figure*}[!tbp]
    \centering
    \includegraphics[width=\textwidth]{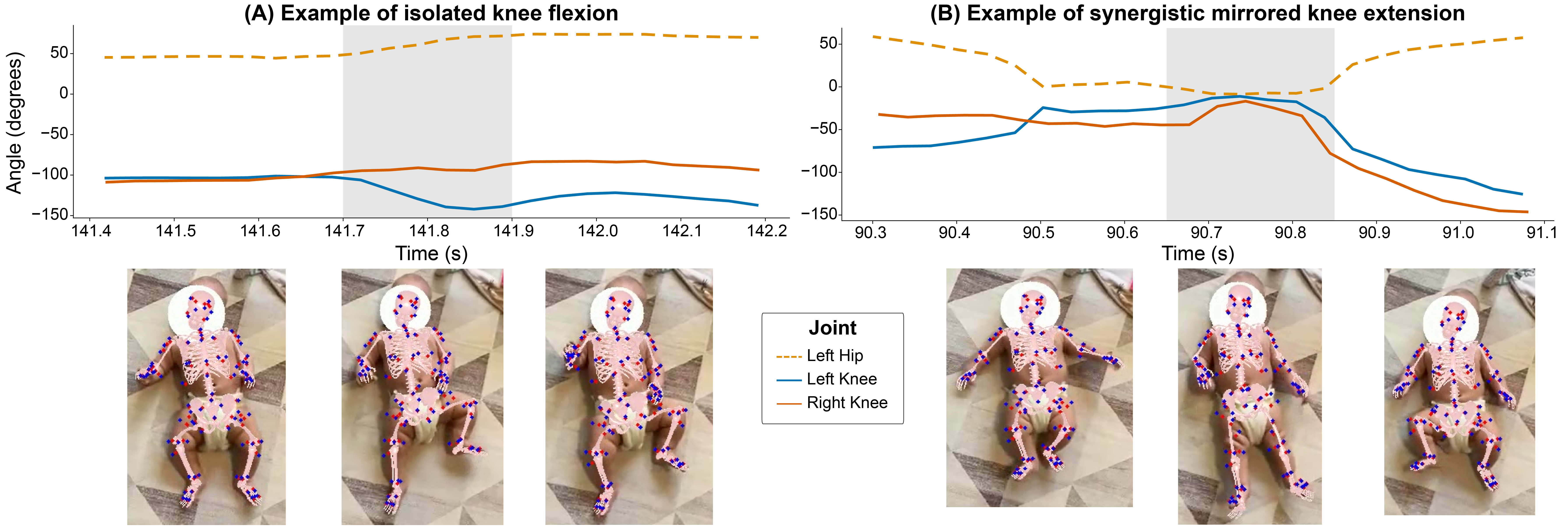}
    \caption{\textbf{Example of monocular kinematic reconstruction from SAM 3D Body (MoVi 87) during representative knee movements in a clinical case example.} Joint angle trajectories are shown during instances of (A) isolated knee flexion and (B) synergistic mirror knee extension. Left hip flexion/extension is shown by a dashed orange line, left knee flexion/extension is shown by a solid blue line, and right knee flexion/extension is shown by a solid red line. Captured frames show overlays of fitted biomechanical model during key moments of the movement. In the frames, detected keypoints are shown in red, model-predicted keypoints are shown in blue, and the fitted model is shown in pink.}
    \label{fig4}
\end{figure*}

We qualitatively evaluated the accuracy of the kinematic reconstructions by generating videos in which the fitted biomechanical model, detected 2D keypoints, reprojected 3D keypoints from the fitted model were overlaid onto the original recordings. We then examined the joint angle trajectories during periods of isolated and patterned movement. Fig. \ref{fig4} illustrates two representative knee movement examples, as identified by an expert clinician, along with the corresponding biomechanical reconstructions and joint angle waveforms.

Fig. \ref{fig4}A shows an instance of isolated knee flexion, in which the left knee moves into flexion while the left hip and the right knee remain relatively stationary. Consistent with this behavior, the corresponding angle–time curves show a clear trough in the left knee angle at the point of maximum flexion, with minimal changes in the other joints. In contrast, Fig. \ref{fig4}B  depicts a case of synergistic mirrored knee extension, where the left knee extends concurrently with the left hip (synergy) and the right knee (mirroring). Accordingly, the joint angle trajectories show simultaneous peaks in the left and right knee angles, accompanied by a decrease in the left hip angle, at the point all three joints reach full extension. Overall, the biomechanical reconstructions, joint angle waveforms, and observed movements in the raw video exhibited close agreement, supporting the potential use of these computer vision algorithms in tracking biomechanical movement of infants from a single video.

\section{Discussion}

This work provides the first comprehensive evaluation of 3D pose estimation in infants and establishes proof-of-concept for biomechanical inverse kinematics during spontaneous infant movements. A key finding is the trade-off between spatial resolution and dimensionality: Sapiens captures fine-grained landmarks but does not yield precise 3D positional estimates, while methods that estimate 3D poses directly (MeTRAbs, SAM 3D Body) were less consistent in landmark resolution. Since dense 3D estimates are essential for fitting biomechanical trajectories, these findings motivate fine-tuning 3D keypoint detectors on large infant datasets to achieve reprojection errors of 5 to 10 pixels, as demonstrated in adults\cite{Cotton2024DifferentiableCapture,Firouzabadi2024BiomechanicalCapture,Cotton2023ImprovedEstimation,Peiffer2025PortableSmartphone}.

For the MoVi 87 keypoints used in kinematic reconstruction, SAM 3D Body matched or outperformed MeTRAbs in position error across all landmarks despite similar reprojection errors, highlighting that these metrics assess distinct aspects of performance. Reprojection error reflects accuracy in camera space, while position error reflects recovery of 3D joint geometry. For biomechanical analysis, where joint configuration matters more than absolute spatial placement, position error is the more relevant metric. SAM 3D Body (MoVi 87)'s consistently lower position error therefore makes it the preferred method for downstream kinematic modeling in infant populations.

As an initial proof-of-concept, we demonstrate that our previously established biomechanical models can be successfully fit to infant data. Using keypoints detected with the SAM 3D Body (MoVi 87) approach, our kinematic reconstruction pipeline reproduced clinician-identified movement classifications, accurately distinguishing isolated from patterned knee movements in a single case example. These preliminary findings suggest that the proposed biomechanical framework can capture clinically meaningful kinematic trajectories in infants, consistent with those commonly evaluated in standard clinical assessments \cite{Peyton2024BabyYears,SukalMoulton2024BabyPalsy,Barbosa2024BabyPerformance}.

Several limitations warrant discussion. First, Sapiens hand keypoints could not be incorporated into kinematic reconstructions despite its superior reprojection error and geometric consistency, as Sapiens detects only 2D keypoints and biomechanical modeling requires 3D pose estimates\cite{Peiffer2025PortableSmartphone}. Nevertheless, SAM 3D Body (MHR 70) hand keypoints closely matched Sapiens estimates, supporting its use for distal joint localization; we therefore plan to integrate these into our pipeline, given the importance of individuated hand movements for characterizing early motor pathway integrity \cite{Peyton2024BabyYears,Novak2017,Cahill-Rowley2014EtiologyPalsy,Spittle2008ALife}. Second, the landmark mapping for SAM 3D Body (MoVi 87) was derived from adult data\cite{Cotton2026MonocularModels}, and may not optimally reflect infant anatomy; future work will learn landmark locations directly from infant data. Third, this pilot was restricted to healthy, term-born one- to four-month-olds, and future studies will evaluate the approach in larger, more diverse cohorts including medically complex infants in neonatal ICU settings.

A further limitation is the lack of ground truth for 3D keypoint validation. Marker-based motion capture introduces soft tissue artifacts and occludes visual pose estimation\cite{MERCIER2026113341}, while fluoroscopy is contraindicated in infants due to ionizing radiation \cite{Strauss2006}. We therefore used multi-view triangulation as a reference standard, which, though imperfect, remains among the strongest available options for markerless infant pose estimation.

Finally, a key future direction is integrating the complementary strengths of both Sapiens and SAM 3D Body, leveraging Sapiens' geometric precision to improve SAM 3D Body's outputs required for biomechanical modeling,  to produce a more accurate and clinically useful movement assessment pipeline.

In summary, this work provides a proof-of-concept that accurate 3D keypoint estimation in infants enables biomechanical model fitting tailored to this population, yielding meaningful insight into joint angle trajectories during critical movement phases. These early findings lay the groundwork for future efforts to refine infant-specific biomechanical frameworks and extend advances in markerless motion capture and biomechanics to the youngest populations.

\section{Conclusion}

Technologies that automatically and non-invasively assess an infant's movement hold great promise for early diagnosis and intervention in infants at risk for movement disorders. As a first step toward this goal, we demonstrate the feasibility of reconstructing full-body joint kinematics from video recordings alone. These preliminary findings open the door to leveraging large-scale video datasets, from NICUs, clinical assessment archives, and even parents' smartphone recordings, to develop predictive models of clinical outcomes based on reconstructed 3D kinematic features

\section*{Acknowledgments}

We would like to to thank Imani Mann and Grace Hoo for their assistance with participant recruitment and scheduling, Shawana Anarwala and Kayan Abdou for their assistance with experimental setup and data collection, and Kunal Shah for his assistance with infrastructure. This work was funded by the National Institutes of Health grants T32TR005124 and 1R21HD117074.

{\small
\printbibliography
}

\end{document}